\documentclass{article}
\usepackage{tcolorbox}
\usepackage{amsmath}
\usepackage{float}
\usepackage{subfig}
\usepackage{algorithm}
\usepackage{algorithmic}
\usepackage[nonatbib]{neurips_2024}
\usepackage{pifont}

\usepackage[utf8]{inputenc} % allow utf-8 input
\usepackage[T1]{fontenc}    % use 8-bit T1 fonts
\usepackage{hyperref}       % hyperlinks
\usepackage{url}            % simple URL typesetting
\usepackage{booktabs}       % professional-quality tables
\usepackage{amsfonts}       % blackboard math symbols
\usepackage{nicefrac}       % compact symbols for 1/2, etc.
\usepackage{microtype}      % microtypography
\usepackage{xcolor}         % colors
\usepackage{enumerate}

\usepackage[draft,textsize=footnotesize,textwidth=15mm]{todonotes}

\title{Density Adaptive Attention is All You Need: Robust Parameter-Efficient Fine-Tuning Across Multiple Modalities}

\author{%
  David S.~Hippocampus\thanks{Use footnote for providing further information
    about author (webpage, alternative address)---\emph{not} for acknowledging
    funding agencies.} \\
  Department of Computer Science\\
  Cranberry-Lemon University\\
  Pittsburgh, PA 15213 \\
  \texttt{hippo@cs.cranberry-lemon.edu} \\
}

\begin{document}

\maketitle

\begin{abstract}
We propose the Multi-Head Density Adaptive Attention Mechanism (DAAM), a novel probabilistic attention framework that can be used for Parameter-Efficient Fine-tuning (PEFT), and the Density Adaptive Transformer (DAT), designed to enhance information aggregation across multiple modalities, including Speech, Text, and Vision. DAAM integrates learnable mean and variance into its attention mechanism, implemented in a multi-head framework, enabling it to collectively model any probability distribution for dynamic recalibration of feature significance. This method demonstrates significant improvements, especially with highly non-stationary data, surpassing the state-of-the-art attention techniques in model performance, up to approximately +20\% (abs.) in accuracy. Empirically, DAAM exhibits superior adaptability and efficacy across a diverse range of tasks, including emotion recognition in speech, image classification, and text classification, thereby establishing its robustness and versatility in handling data across multiple modalities. Furthermore, we introduce the Importance Factor, a new learning-based metric that enhances the explainability of models trained with DAAM-based methods\footnote{\url{https://github.com/gioannides/DAAM-PEFT-paper-code}}.
\end{abstract}

%\section{Introduction}
Attention mechanisms, as exemplified in the Transformer model \cite{vaswani2017attention}, have significantly advanced the field of sequence modeling, particularly in Natural Language Processing (NLP) and various branches of Signal Processing such as Speech Processing and Digital Image Processing. These mechanisms are adept at capturing dependencies within the context length, although their effectiveness can vary based on the relative placement of tokens and the inherent limitations in handling long-range dependencies due to quadratic complexity \cite{wang2023rrwkv, zhuang2022long, he2023long}. Ongoing research continues to address these challenges, seeking more efficient ways to model long sequences and capture global context dependencies.

In recent years, the self-attention mechanism, or its variations that are based on dot-product, has become central to the encoders of Pre-Trained Models (PTMs) developed using Self-Supervised Learning (SSL) methods. Notable examples include WavLM \cite{9814838} and HuBERT \cite{hsu2021hubert} for speech processing, Llama 2 \cite{touvron2023llama} for text processing, and BEiT \cite{bao2022beit} for image processing. These PTMs are highly effective at generating contextualized embeddings, outperforming traditional feature engineering methods \cite{fearless}, and they are adaptable to a wide range of downstream tasks specific to their training modalities.

Section \ref{relatedwork} reviews relevant previous work and explains DAAM and its integration with dot-product attention (Grouped Query Attention); Section \ref{results} discusses experimental findings, limitations and future research; Section \ref{conclusion} concludes. Our contributions can be summarized as follows:

\begin{enumerate}
    \item \textls[-3]{We propose the Multi-Head DAAM and the Density Adaptive Transformer (DAT), featuring fully learnable mean and variance parameters within a multi-headed parameter-efficient framework. This enables dynamic recalibration of feature importance to fit any probability distribution best suited to attend the data it is trained on.}
    
    \item Introduction of IF, a new metric to enhance explainability in models using DAAM, quantitatively assessing feature significance for improved interpretability.
    
    \item Utilization of PTMs as an embedding extractors across Speech, Text, and Vision modalities, demonstrating DAAM's superiority over conventional dot-product attention and other Parameter-Efficient Fine-tuning methods in handling non-stationary data.
    
    \item Integration of DAAM with Grouped Query Attention, optimizing computational efficiency and showcasing compatibility with dot-product attention in popular PTM models (e.g., WavLM, HuBERT, Llama), enhancing performance with minimal parameter increase.
\end{enumerate}

\section{Methods}
\label{relatedwork}

\subsection*{Advantages of Learning with Multi-Head DAAM}
DAAM leverages additive and multiplicative Gaussian parameters -- mean offsets and variance scaling factors, respectively -- to dynamically adjust attention. The mean offset shifts the Gaussian's focus based on input context, enhancing responsiveness to deviations. In parallel, the variance scaling adapts the distribution's spread, ensuring attention is not only accurately centered but also suitably scaled to the task's specificity. The multi-head design allows each head to address different data aspects, enhancing the model's adaptability to non-Gaussian traits \cite{Fluri2022}.

Below, we analyze the theoretical aspects of entropy for both traditional self-attention mechanisms and DAAM.

\subsubsection*{Density Adaptive Attention with $N$ Gaussians per Head, $h$} 

Each head, \(h\), in Density Adaptive Attention Mechanism (DAAM) processes input using Gaussian normalization, which is controlled by learnable parameters \(\mu_{i,h}\) and \(\sigma_{i,h}\). The transformation is defined by the formula \( y_{\text{norm}} = \frac{y - (\text{mean} + \text{mean\_offset})}{\sqrt{\text{var} + \epsilon}} \), where \(\epsilon\) is a small constant ensuring numerical stability. This normalized input is then applied to a Gaussian function, \( f^{(h)}(x) = \exp\left(-\frac{y_{\text{norm}}^2}{2c^2}\right) \), with \(c\) as a learnable parameter that controls the spread of the Gaussian function. The overall transformation for each head approximates a Gaussian distribution, where the variance \(\sigma_{\text{prod}}^2\) is a function of the aggregated variances and mean adjustments within the head, represented by \( \sigma_{\text{prod}}^2 = \left( \sum_{i=1}^N \frac{1}{\sigma_{i,h}^2} \right)^{-1} \). Additionally, the effective mean is given by \( \mu_{\text{prod}} = \sigma_{\text{prod}}^2 \left(\sum_{i=1}^N \frac{\mu_{i,h}}{\sigma_{i,h}^2}\right) \). 

The entropy for each head, denoted as \( H(X_h) \), is calculated using the formula \( \frac{1}{2} \log(2\pi e \sigma_{\text{prod}}^2) \). This entropy value reflects how the data is spread, influenced by parameters such as \(c\), the mean offset, and the computed variance from the downsampled data. To capture the overall system entropy, including potential interactions among multiple heads, it is represented by the formula \( H(\text{X}) = \sum_{h=1}^H H(X_h) + \Delta \). Here, \(\Delta\) accounts for additional entropy arising from the diversity and interactions across different heads, highlighting the ensemble effect of the multi-head Gaussian transformations. This approach allows DAAM to modulate attention distribution adaptively, balancing between broad and focused attention based on input characteristics.

\subsubsection*{Traditional Self-Attention} 

Traditional self-attention mechanisms are mathematically represented as \( \text{Attention}(Q, K, V) = \text{softmax}\left(\frac{QK^T}{\sqrt{d_k}}\right)V \). In this framework, the softmax function is applied to the scaled dot products of queries (Q) and keys (K), producing attention weights. Consider a vector \( z = \{z_1, z_2, \ldots, z_n\} \), derived from these scaled dot products. Let \( S = \sum_{j=1}^n e^{z_j} \) denote the sum of exponential terms. The softmax values are represented as \( \left\{\frac{e^{z_1}}{S}, \frac{e^{z_2}}{S}, \ldots, \frac{e^{z_n}}{S} \right\} \), with entropy \( H(\text{softmax}(z)) = -\sum_{i=1}^n \left( \frac{e^{z_i}}{S} \log \frac{e^{z_i}}{S} \right) \). Typically, this entropy is low unless the \(z\) values are nearly identical, leading to a uniform softmax output. This low entropy results from the exponential nature of the softmax function, which tends to emphasize larger dot product values, thereby focusing attention on specific parts of the input.

In practical terms, the output of traditional self-attention often skews towards dominant features, concentrating attention and leading to lower entropy. Higher entropy, indicating a more uniform attention distribution, can be beneficial for tasks requiring a comprehensive view of all input data. Achieving higher entropy theoretically demands near-uniformity in the elements of \( z \) (the inputs to the softmax function). However, without modifications to the architecture—such as designing or constraining the weight matrices \( W^Q \) and \( W^K \) to produce similar outputs across different inputs—traditional self-attention mechanisms inherently produce lower entropy. This characteristic makes them less adaptable in scenarios demanding sensitivity to diverse and dynamic data elements, highlighting a limitation in their design for certain applications.

Higher entropy in an attention mechanism signifies a more balanced distribution of attention across various parts of the input data. This balance is crucial for ensuring that the model does not overly focus on a few features but instead considers a broader array of information. Such capability is vital in complex tasks where the input data is highly non-stationary. \textbf{In contrast to traditional self-attention, the DAAM architecture dynamically adjusts its entropy in response to input characteristics. It provides both broad (high entropy) and focused (low entropy) attention distributions as needed, which is essential for effectively handling both highly non-stationary and stationary data environments.}

\vspace{-2mm}
\subsection{Attention Mechanisms \& Parameter Efficient Fine-Tuning}

The Multi-Head Attention (MHA) mechanism in Transformer architectures uses parallel attention heads to enhance sequence modeling. Each head computes attention scores independently using the scaled dot-product attention formula:

\[
\text{Attention}(Q, K, V) = \text{softmax}\left(\frac{QK^T}{\sqrt{d_k}}\right)V,
\]

where \(d_k\) is the dimensionality of the keys. This enables each attention head to focus on different parts of the input, capturing diverse relationships within the data. The overall MHA is expressed as:

\[
\text{MHA}(Q, K, V) = \text{Concat}(\text{head}_1, \ldots, \text{head}_H)W^O,
\]

with each head calculated as:

\[
\text{head} = \text{Attention}(QW_i^Q, KW_i^K, VW_i^V).
\]

This architecture allows the model to capture complex patterns and dependencies, improving performance in various tasks, such as machine translation and text summarization.

The Grouped Query Attention (GQA) mechanism serves as an intermediary between MHA and Multi-Query Attention (MQA). In GQA, query heads are grouped into \(G\) groups, with each group sharing a single key and value. This can be formulated as:

\[
\text{GQA}(Q, K, V, G) = \text{Concat}(\text{group}_1, \ldots, \text{group}_G)W^O.
\]

GQA provides a balance between computational efficiency and the model's capacity to learn complex relationships, making it a valuable tool for scaling attention mechanisms in large models. 

Low-Rank Matrix Adaptation (LoRA) is a technique designed for efficient model adaptation, commonly used in the Natural Language Processing domain. LoRA leverages low-rank decomposition to fine-tune pre-trained models by introducing low-rank updates to the weight matrices in neural networks. This is parameterized as \(\Delta W = AB^T\), where \(A \in \mathbb{R}^{d \times r}\) and \(B \in \mathbb{R}^{d \times r}\) with \(r \ll d\). This approach significantly reduces the number of parameters while maintaining the model's expressive power. The overall weight update in the model is represented as \(W' = W + \Delta W = W + AB^T\), where \(W\) is the original weight matrix.

By adjusting the rank \(r\), LoRA balances between model complexity and computational efficiency, proving particularly effective for scenarios requiring rapid adaptation to new tasks or domains. This method has demonstrated competitive performance with significantly reduced resource requirements, as noted in recent research~\cite{hu2021lora, ding2022lora, li2021lora}.

Additionally, LoRA+ extends the original LoRA technique by introducing separate learning rates for the matrices \(A\) and \(B\), thereby allowing more fine-grained control over the model adaptation process. This modification enhances the flexibility of LoRA by decoupling the optimization dynamics of the two matrices. Specifically, instead of a shared learning rate for both \(A\) and \(B\), LoRA+ applies different learning rates, \( \eta_A \) and \( \eta_B \), to each matrix during the training process, enabling improved convergence behavior and potentially better adaptation to the target task. This innovation has been explored in more recent literature~\cite{hayou2024loraefficientlowrank}, demonstrating that separate learning rates can further optimize performance while maintaining LoRA's benefits of parameter efficiency and adaptability.

\textbf{Despite these efficiencies, LoRA still requires a relatively higher number of parameters—at least 24 times more than DAAM across the models used and at least 2 times for than DAT in this study, even when assuming the lowest rank possible, \(r=4\). This parameter difference highlights the efficiency of DAAM and DAT in comparison, as illustrated in Table} \ref{tab:attention_comparison}.

\subsection{General Purpose Pre-Trained Models}
WavLM \cite{9814838} is a large-scale pre-trained model designed to enhance speech processing capabilities by utilizing 94,000 hours of diverse audio inputs. Building upon the Hidden Unit BERT (HuBERT) framework \cite{hsu2021hubert}, which primarily focuses on masked speech prediction, WavLM incorporates an additional denoising modeling component. This dual approach allows WavLM to handle a broader range of speech-related tasks effectively. At its core, WavLM utilizes self-supervised learning techniques, enabling the model to predict masked portions of audio inputs. Through this process, the model acquires a deeper understanding of speech patterns, nuances, and contextual information, thereby improving its performance in various speech processing applications.

Llama family of models \cite{touvron2023llama} mark a significant advancement in Large Language Models (LLMs), leveraging an optimized auto-regressive transformer architecture. %One of its key enhancements is training on a token volume of 2 trillion tokens sourced from publicly available data. Llama 2 also doubles the context length to 4096 tokens compared to its predecessors, which significantly improves its capability to comprehend and generate longer text sequences. The integration of Grouped-Query Attention (GQA) further enhances its scalability, particularly when applied to larger models, allowing for more efficient inference.

Bidirectional Encoder Representations from Image Transformers (BEiT) \cite{bao2022beit} represents a breakthrough in self-supervised learning for vision tasks, drawing inspiration from the BERT \cite{Devlin2019BERTPO} approach in natural language processing. BEiT utilizes Masked Image Modeling (MIM) as a pre-training strategy for vision transformers. In this approach, images are tokenized into discrete visual tokens, and a blockwise masking strategy is applied. The model then predicts the original visual tokens from these masked patches, focusing on learning high-level semantic representations directly from raw pixel data. This methodology allows BEiT to excel in downstream tasks such as image classification and semantic segmentation, surpassing other pre-training methods in both performance and fine-tuning stability.
%\vspace{-0.6in}
%\section{Methods}
\begin{figure}[ht]
\centering
\includegraphics[width=1.0\textwidth]{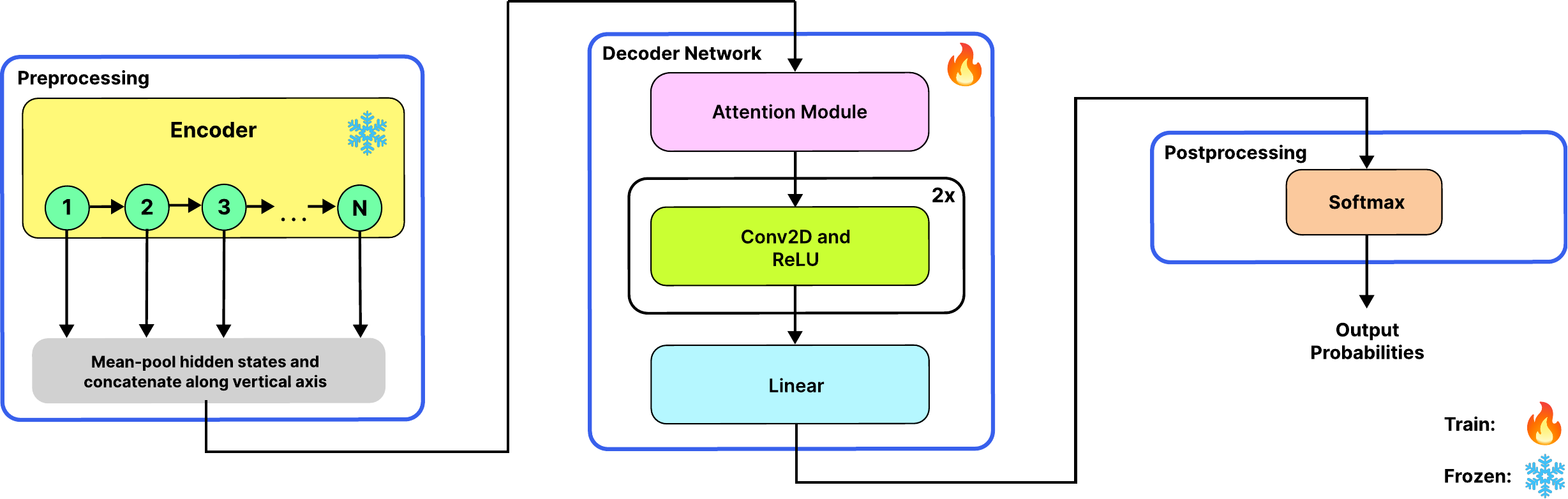}
  \caption{Proposed model architecture showcasing a pre-trained model (i.e., the encoder) for feature extraction (i.e., embeddings) via its $N$ transformer layers, followed by the attention module within the decoder network for selective emphasis, and concluding with probability output. The process flow is marked with the trainable and frozen states.}
  \label{model-arch}
\end{figure}
\subsection{Datasets}
\label{datasets}
The frozen encoder of each of the three PTM implementations (as described in Section \ref{relatedwork}) is used to train and evaluate a decoder on the \textit{IEMOCAP} \cite{iemocap}, \textit{AG News} \cite{agnews} and \textit{CIFAR100} \cite{cifar100} datasets to assess the applicability of the newly proposed attention mechanisms across speech, text and image modalities. For the \textit{IEMOCAP} dataset, we employ 5-fold cross-validation, training on 4 sessions and validating on 1. We focus on the emotion categories \textit{neutral}, \textit{happiness} (merging \textit{happiness} and \textit{excited}), \textit{anger}, and \textit{sadness}. \textit{AG News} dataset is employed in our study. Our dataset construction focuses solely on the title and description fields of these articles. In terms of data distribution, each category (out of four) contributed 30,000 articles to the training set and 1,900 articles to the validation set. For our analysis, we use the following division of the CIFAR-100 dataset: 50,000 images for training and 10,000 for validation.
%\vspace{-10mm}
In the Multi-head DAAM, as outlined in Algorithm \ref{alg:gaussian_adaptive_attention}, the attention weights are determined utilizing multiple Gaussian probability density functions. In this formulation, the scaled variances are treated as learnable parameters, while the means are adjusted through learnable offsets. This methodology enables the model to dynamically adapt its attention focus in response to the distribution of the input data.

In a multi-head configuration, the DAAM process is independently applied across different heads, each focusing on distinct subspaces of the original input features. The final output (of all heads combined via stacking) is computed as an element-wise multiplication (Hadamard product) of the original input features and the Density attention weights.

This process enhances the model's ability to focus on contextually relevant information in the input sequence. All head outputs are stacked \textit{vertically}, forming the Density Attention modulated Tensor ($X'$). An even more parameter-efficient solution of the DAAM -- termed as the \textit{Mixture of Densities Adaptive Attention Mechanism}, can be found in the Appendix (see Section \ref{optimized}). 
Following the integration of Multi-Head DAAM, we investigate its compatibility with dot-product-based attention mechanisms (e.g., MHA, MQA, and GQA). Our focus on GQA is driven by its comparable performance to MHA, superior computational efficiency \cite{gqa} and advantages of its hierarchical learning structure \cite{hierarchy}. We refer to this method as the Grouped Query Density Adaptive Attention Mechanism (GQDAAM). The objective here is to showcase that DAAM can benefit PTMs across multiple modalities as a parameter-efficient fine-tuning method. The Computational complexity of DAAM can be analyzed as follows: $O(n \cdot m)$, where $n$ is the batch size and $m$ is the dimension size along \textit{normAxis}. $O(h \cdot n \cdot m)$, with $h$ as \textit{numHeads}, allowing for parallelization.

\begin{algorithm}
\caption{Density Adaptive Attention Mechanism}
\label{alg:gaussian_adaptive_attention}
\begin{algorithmic}[1]
\REQUIRE $x$ (input tensor), normDimSize, normAxis, c, eps
\ENSURE Attention-modified tensor
\STATE Initialize $\mathbf{c}$ to a tensor of size $(1, \text{normDimSize})$ filled with $c$
\STATE Initialize $\mathbf{meanOffset}$ to a tensor of size $(1, \text{normDimSize})$ filled with zeros
\FOR{each batch in $x$}
    \STATE $\mathbf{mean} \gets \text{mean}(x, \text{dim}=\text{normAxis})$
    \STATE $\mathbf{var} \gets \text{mean}(x^2, \text{dim}=\text{normAxis}) - \mathbf{mean}^2$
    \STATE $\mathbf{var} \gets |\mathbf{var}| + 10^{-8}$
    \STATE $\mathbf{adjustedMean} \gets \mathbf{mean} + \mathbf{meanOffset}$
    \STATE $\mathbf{yNorm} \gets (x - \mathbf{adjustedMean}) / \sqrt{\mathbf{var} + 10^{-5}}$
    \STATE $\mathbf{yTransform} \gets \exp(- (\mathbf{yNorm}^2 / (2 \cdot \mathbf{c})))$
    \STATE $x \gets x \cdot \mathbf{yTransform}$
\ENDFOR
\RETURN $x$
\end{algorithmic}
\end{algorithm}

\subsection{Encoder and Decoder Models}
We apply different attention mechanisms on SSL-based PTMs acting as PTMs to extract embeddings from. Specifically, we utilize the pre-trained model weights from three distinct PTMs: (i) WavLM-Large, (ii) Llama2-13B, and (iii) BEiT-Large.

Libri-light \cite{librilight} GigaSpeech \cite{GigaSpeech2021} and English parts of VoxPopuli \cite{voxpouli} have been used for pre-training (i). ImageNet-1k \cite{Russakovsky2015} has been used for pre-training (iii). Conversely, while (ii) has undergone pre-training on undisclosed, publicly sourced textual data, this aspect does not impact our research. The downstream application we employ is different from the model's original pre-training task. The specific datasets used in this work are described in further detail in Section \ref{datasets}. 

The role of PTMs within the proposed model architecture, as depicted in Figure \ref{model-arch}, is crucial during the inference phase (post-training). It is important to note that in this study, PTMs are utilized in their original pre-trained state, eschewing any further re-training during the preprocessing stage. For each PTM under consideration, the encoder component remains static (frozen), allowing the focus to be on training and subsequently evaluating the performance of the newly proposed decoder on the designated downstream task. This approach ensures that the intrinsic properties and learned representations of the PTMs are preserved, while the decoder adapts and fine-tunes to the specific requirements of the task at hand \cite{ioannides23_interspeech}.
The output from each transformer layer (in the encoder) undergoes mean pooling across the time dimension (sequence length), followed by concatenation of these pooled outputs. These concatenated outputs then serve as input embeddings for the Attention Module, which employs either (i) Multi-Head Self-Attention, (ii) Multi-Head Density Adaptive Attention, or (iii) Multi-Head Grouped Query Density Adaptive Attention where (ii) and (iii) are contributions of this work. \textbf{When (ii) or (iii) are used, the decoder network is termed as DAT}.

In mathematical terms, the embeddings are represented as \(X \in \mathbb{R}^{N \times d}\), where each \(x_i\) is a vector in a \(d\)-dimensional space, with \(d\) taking values in the set \(\{1024, 5120\}\). Here, \(N\) signifies the total count of transformer layers in the encoder, which are kept in a static (frozen) state. The attention mechanism of the module then produces a new, contextualized representation \(C \in \mathbb{R}^{N \times d}\) for the input sequence. 
Subsequently, convolutional layers are utilized to distill features (pertaining to speech, text, or image data) from the context matrix generated by the attention mechanism. By employing 2-dimensional convolution layers (with $\text{kernel\_size}=(3,3)$, stride = 1, and padding = 1), the model adeptly processes the array of context tensor outputs from each transformer layer. 

Table \ref{tab:attention_comparison} lists the attention mechanism parameters for the proposed DAAM-based decoders of WavLM-Large, Llama2-13B, and BEiT-Large encoders. Here, $g$ denotes DAAM-based head count, with higher values indicating a higher number of learned Gaussian Distributions. $q$ and $kv$ are the counts of query and key-value heads, respectively. The embedding dimensions are 1024 for WavLM and BEiT, and 5120 for Llama2. We also benchmark against LoRA-based methods downstream task by fine-tuning the query and key projections modules.
\begin{table}[t!]
\centering
\footnotesize % Smaller font size for the table content
\setlength\tabcolsep{4pt} % Reduces the space between columns
\begin{tabular}{lccc}
\toprule
\textbf{Mechanism} & \textbf{Heads} & \textbf{$N$ Gaussians} & \textbf{Parameters (Millions)} \\
\midrule
GQDAAM & \(g: 8, q: 8, kv: 2\) & 1024 - 5120 & 1.21 - 3.55 (DAAM -- 0.016 - 0.082) \\
GQA & \(q: 8, kv: 2\) & N/A & 1.19 - 3.47 \\
LoRA+ ($r=\{4,8\}$, $\alpha=16$) & N/A & N/A & 0.39 - 3.28 \\
LoRA ($r=\{4,8\}$, $\alpha=16$) & N/A & N/A & 0.39 - 3.28 \\
DAAMv1 (with 2 \textit{conv}. layers) & \(g: 8\) & 1024 - 5120 & 0.22 - 0.45 (DAAM -- 0.016 - 0.082) \\%0.016 - 0.082 \\
DAAMv2 (with 2 \textit{conv}. layers) & \(g: 1\) & 1024 - 5120 & 0.22 - 0.45 (DAAM -- 0.002 - 0.010) \\
\bottomrule
\end{tabular}
\vspace{2mm}
\caption{Comparison of min and max learnable parameters (in millions) for various PEFT methods.}
\label{tab:attention_comparison}
\vspace{-2mm}
\end{table}

All decoder network models are trained for 35 epochs and their layer weights (except their respective attention module) are initialized using Xavier initialization \cite{xavierinit}. Adam \cite{kingma2017adam} is used as the optimizer, with both weight decay factor of $0.1$ and an initial learning rate of $10^{-4}$ (except for when Llama 2 is used as an encoder, in which case it is $5 \times 10^{-5}$). For the SER experiments, Focal Loss \cite{lin2018focal} is used, where $\gamma=2.5$. For the text and image classification experiments the Cross-Entropy Loss is used. All DAAM-based attention modules are initialized with a mean offset, $\delta = 0$, where $\delta \in 1 \times d$ and scaled variance, $\xi = 2$, where $\xi \in 1 \times d$. A batch size of 8 is used for SER and a batch size of 32 for Text and Image Classification. Across all downstream tasks, mixed precision training is utilized. Regarding the SER downstream task -- during training and evaluation, audio files are split to a maximum of 5 second clips. If an audio file exceeds 5 seconds in duration, a new audio file will be generated containing the excess audio. Each audio file is passed through the trained Encoder model. For the text classification downstream task, text is tokenized with maximum context length of 4096 during both training and evaluation. For the image classification downstream task, images are resized to $\text{224} \times \text{224}$ during both training and evaluation.
\subsection{Evaluation Metrics}
In this study, the primary evaluation metric is Accuracy (Acc.), calculated as the percentage of correct predictions to total predictions. Additionally, the Importance Factor (IF) is introduced, calculated using Density Attention weights (DA) from the attention module. IF is $\frac{\text{DA}_{ij} - \min(\text{DA})}{\max(\text{DA}) - \min(\text{DA})}$ with $\text{IF} \in [0, 1]$, indicating the relative importance of features in the model's decision process. Higher IF values indicate more significant features and vice versa. IF-based heatmaps are created by taking the arithmetic average of the generated Density Attention maps during validation and then applying the $\text{IF}$ formula. They visually depict feature importance. Unlike traditional self-attention, where attention might skew towards a few dominant features, DAAM's Gaussian-based attention provides a more balanced spread. This helps in capturing a broader range of features, reducing the bias towards overly frequent features and focusing more on features that contribute meaningfully to the task. All experiments have been carried out on two A100 80GB NVIDIA Graphical Processing Units (GPUs).

\section{Discussion}
The motivation for this research arises from the broad range of downstream applications that could benefit from an improved attention mechanism, addressing the limitations inherent in the self-attention mechanism (or other dot-product attention mechanism variations) found in Transformer models, which rely on normalized dot-products. This presents an opportunity to explore more robust and explainable approaches.

Despite their widespread adoption, self-attention mechanisms face several limitations that can impact their performance. One significant challenge is the fixed-length context window, which can lead to sub-optimal outcomes, particularly for long sequences where distant elements may lack relevance \cite{li2023unlocking}. Additionally, without inductive biases like locality \cite{edelman2021inductive}, self-attention layers may require more data to learn patterns that other methods can capture more efficiently. Although self-attention is theoretically capable of capturing long-term dependencies, it can struggle in practice as sequence length increases \cite{10.1162/tacl_a_00306}. Furthermore, the interpretability of self-attention mechanisms is limited; it primarily relies on correlation-based activations, which focus on pairwise similarities and may not effectively capture the most relevant context \cite{xai}. This makes it difficult to understand why certain parts of the input are prioritized, underscoring the need for more transparent and interpretable attention mechanisms/frameworks.

In our work, we introduce a significant enhancement to the Transformer model's attention mechanism: the (Multi-Head) Density Adaptive Attention Mechanism (DAAM). DAAM is designed to improve upon the standard self-attention mechanism in Transformers. Unlike conventional attention in the Transformer, which calculates weights based on dot-product between different weight matrices, DAAM employs a Gaussian-based modulation of input features instead. This approach enables the model to concentrate on the most pertinent features in a context-sensitive manner, thereby improving its capability to interpret and process sequential and spatial data. 

DAAM's attention mechanism, can be applied in various domains like multimedia recommendation (as in \cite{tao2022self}), image classification (aligning with Patrick et al.'s \cite{patrick2022reconstructive} robustness strategies), and text classification (enhancing accuracy in contexts like e-commerce as shown by Yıldırım et al. \cite{yildirim2019realworld}), and can significantly enhance model performance. Its ability to dynamically recalibrate feature significance based on Gaussian parameters proves particularly beneficial, offering improved accuracy, robustness, and user experience across diverse and challenging real-world applications. Furthermore, DAAM's Gaussian-based modulation offers a more interpretable framework for Artificial Intelligence (AI), addressing the critical need for transparency and trustworthiness in real-world AI systems \cite{jin2023rethinking}.

Our proposed DAAM mechanism learns multiple means and variances of input features in a Multi-Headed setting. This mechanism operates independently across different heads, each focusing on distinct, non-overlapping subspaces of the input features. By employing Gaussian modulation, DAAM assigns varying levels of importance to each feature, effectively generating local attention outputs from each head. These outputs are then combined to construct a comprehensive Global Attention map. Each head independently adjusts its means and variances, allowing for a focused approach to different skewness aspects in data subsets capturing a broader range of data characteristics, including asymmetries, and collectively, non-Gaussian traits. Unlike other approaches in the literature wherein no parameters of the Gaussian distribution are learned, and are thus hard-coded making them non-specific to the data they are used on \cite{You2020HardCodedGA, Guo_Zhang_Liu_2019}, only multiplicative parameters like the scaled variance are learned \cite{gct, 9053591}, a pre-defined amplitude that is updated during training \cite{Luo_2023_ICCV} which are all approaches that are limited because their attention framework can only explicitly model Gaussian traits behavior \cite{CHEN2023109467, 9511096}.
\section{Results}
\label{results}
\subsection{Current Benchmarks \& Analysis of Encoder Layer Contribution}
\begin{figure*}[ht!]
\centering
% First image (a)
\subfloat[Speech Processing with \textit{WavLM-Large} using DAAM.\label{iemocap_gaam}]{
  \includegraphics[width=0.6\textwidth]{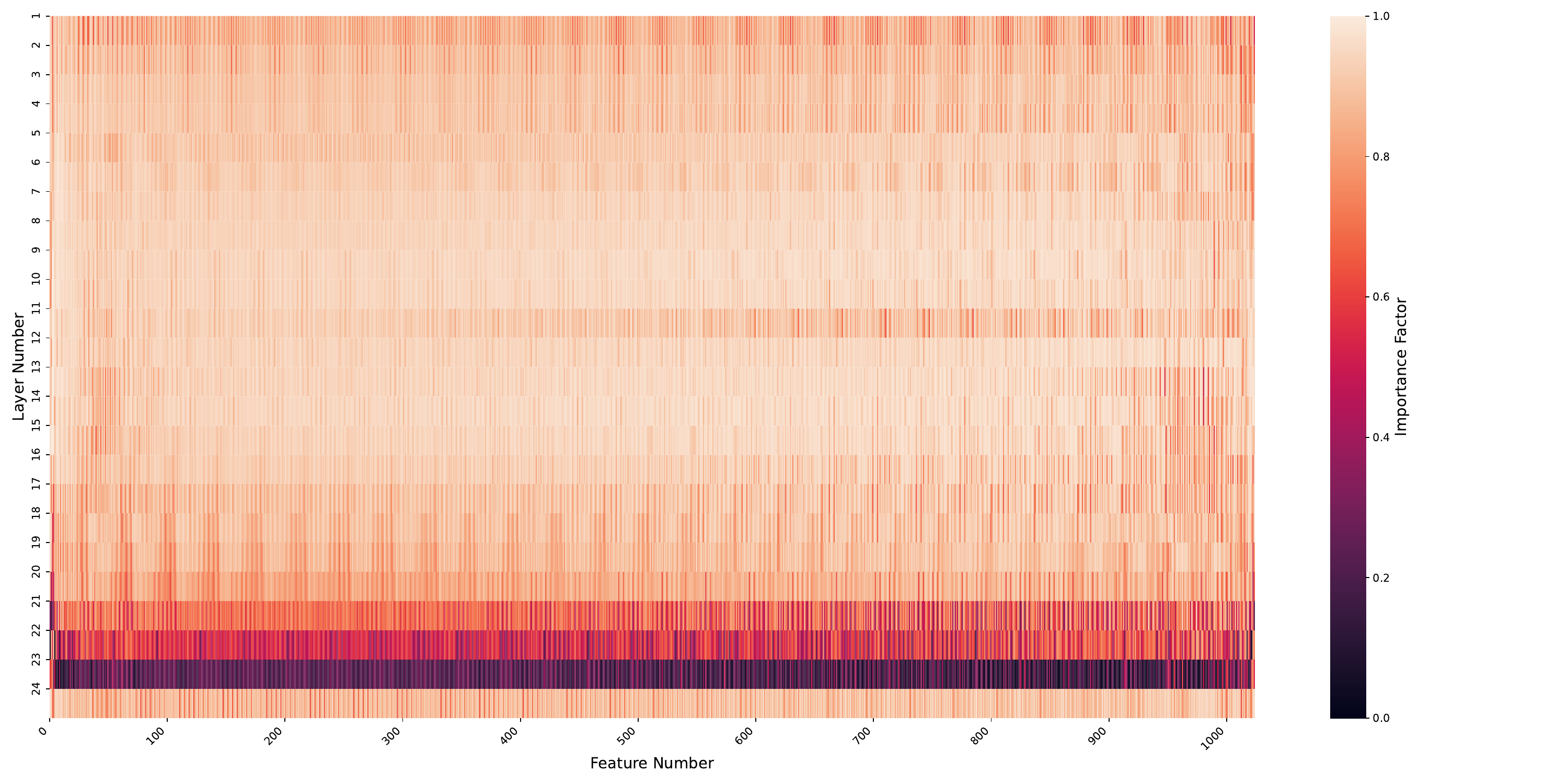}
}

\vspace{0.5cm} % Vertical space between images

% Second image (b)
\subfloat[Text Processing with \textit{Llama2-13B} using DAAM.\label{llama2_gqgaam}]{
  \includegraphics[width=0.6\textwidth]{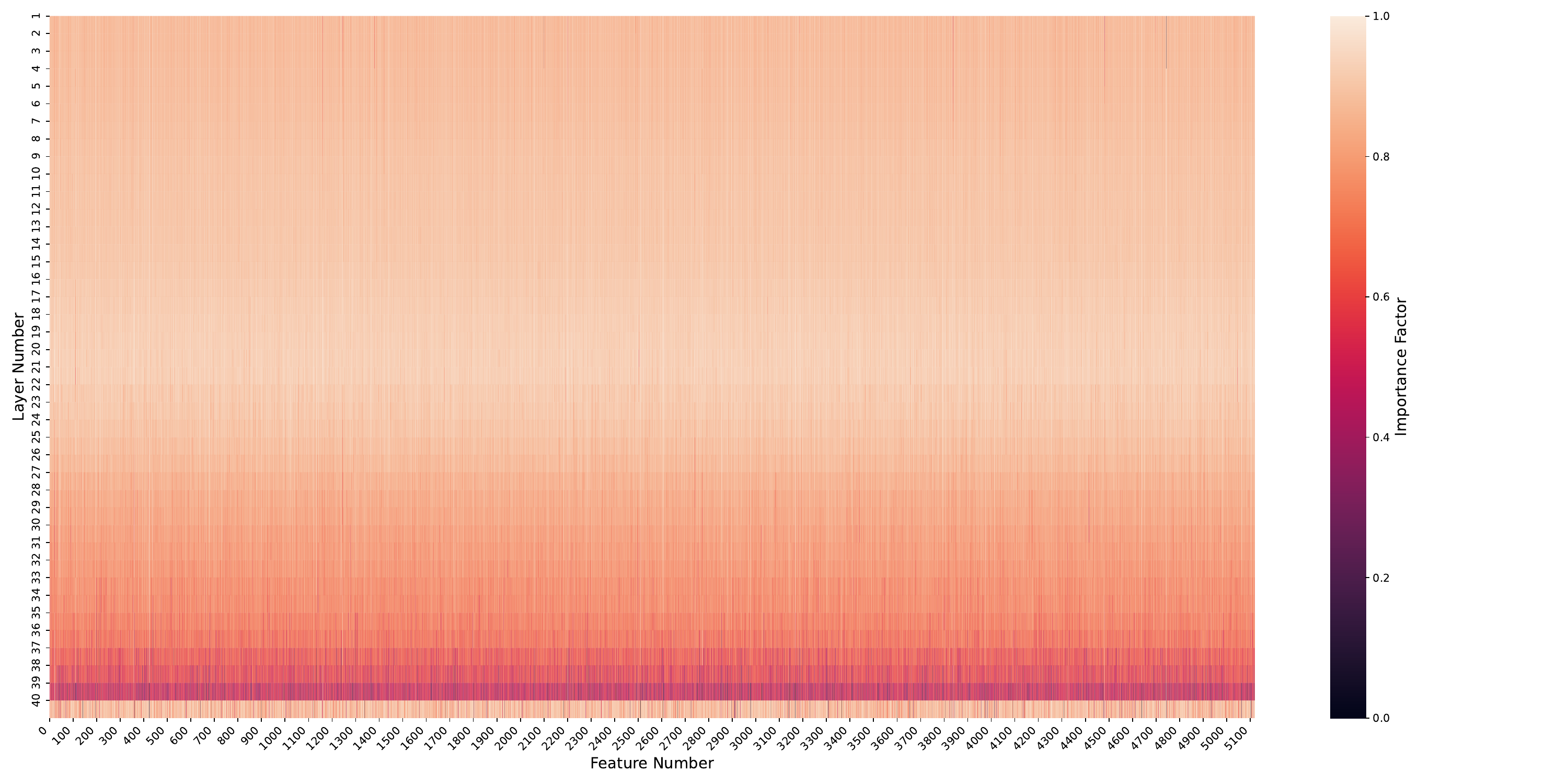}
}

\vspace{0.5cm} % Vertical space between images

% Third image (c)
\subfloat[Digital Image Processing with \textit{BEiT-Large} using DAAM.\label{cifar100_gqgaam}]{
  \includegraphics[width=0.6\textwidth]{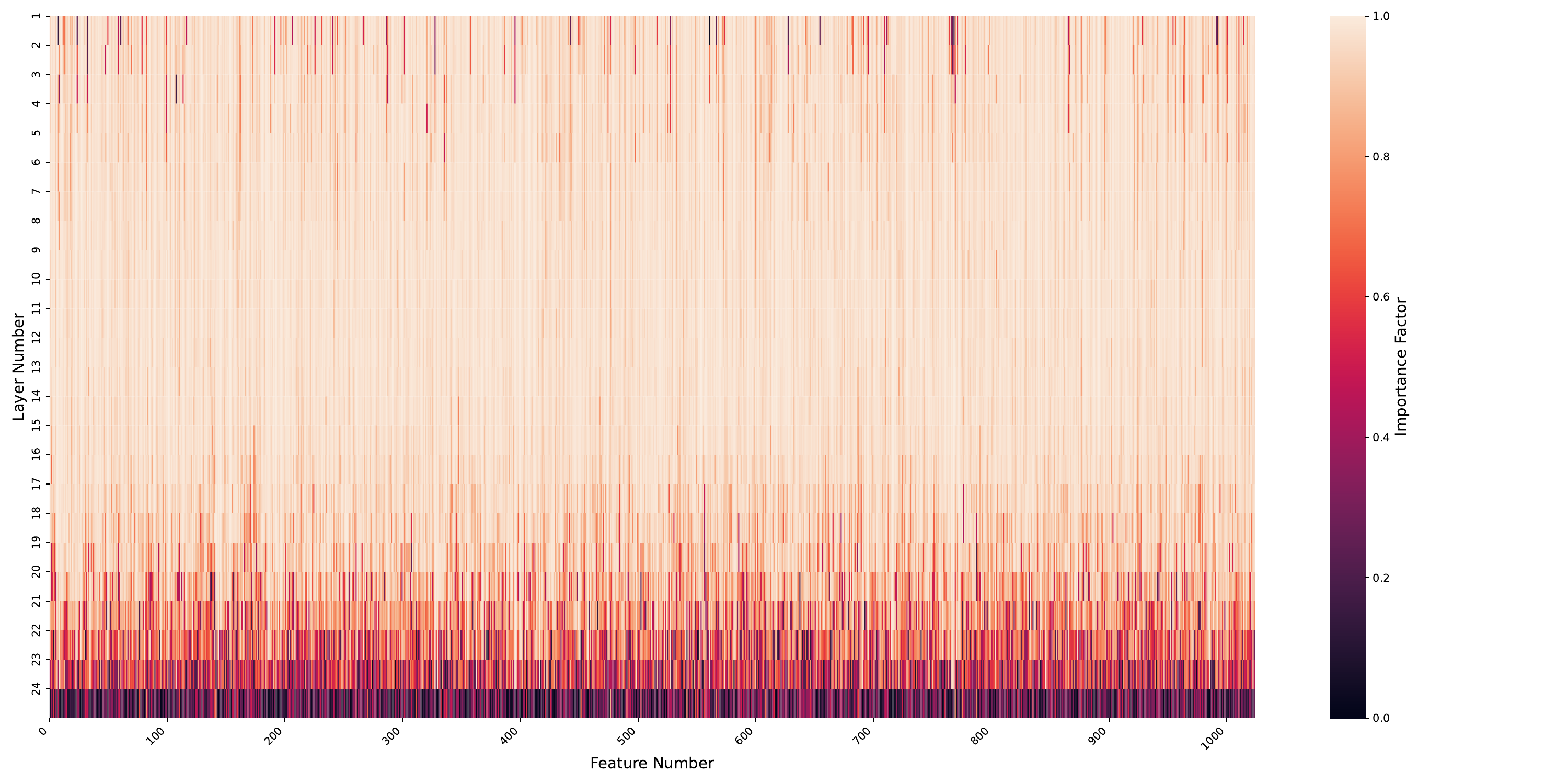}
}

\caption{IF values for different processing tasks with their respective models (with output feature number on the X-axis and layer number on the Y-axis).}
\label{fig:importance_factors}
\end{figure*}

In Table \ref{tab:combined}, we compare DAAM-based methods with those utilizing Multi-Head Self-Attention with and without Batch Normalization (BN); LoRA-based methods (rank of 4 and 8).
BN assumes independent and identical distribution across mini-batches \cite{10095399} and is applied immediately after MHA to maintain the normalization order consistent with the original Transformer architecture. DAAM, due to its multi-head structure, effectively handles variations in feature distribution, including shifts in mean offset and variance, as well as the ability to model a mixture of density distributions with varying attention weights. This capability is demonstrated in Table \ref{tab:combined} and Figure \ref{fig:importance_factors}. Such adaptability enables DAAM to outperform methods dependent on static feature distributions, such as BN.
Unlike methods that assume data follows a single Gaussian distribution, DAAM can model multiple Gaussian distributions with varying parameters, allowing it to approximate any probability distribution. The results indicate that DAAM demonstrates superior performance in scenarios characterized by significant variations in Gaussian parameters, which suggests a need for modeling non-stationary data and potentially non-Gaussian characteristics. Conversely, in cases where such parameter variations are minimal, DAAM outperforms MHA. 

Observing the data in Table \ref{tab:combined}, we note that speech data exhibits high variability in both central tendency (mean offset, \(\mu\)) and spread (scaled variance, \(\sigma^2\)). This variability reflects the highly non-stationary nature of speech \cite{9440639}, necessitating attention mechanisms that can dynamically adjust both focus (\(\mu\)) and width (\(\sigma\)) to capture the rapidly changing features essential for tasks such as emotion recognition. Text data, on the other hand, shows high mean variation, possibly due to changing semantic contexts, while the variance remains moderately stable, which aligns with the structured nature of language. In text processing, attention mechanisms need to focus primarily on tracking the shifting mean to match the changing semantic and syntactic focal points. In contrast, vision tasks show low variations in both mean and variance, indicating that features are relatively stable and consistent in their locations and spreads. This stability suggests that simpler attention mechanisms can be effective, maintaining a consistent focus and width suitable for tasks with minimal feature variability.

To understand the contribution of encoder layers, we employ heatmap visualizations of the Importance Factor (IF) to reveal how features within the frozen pre-trained models drive decision-making. We analyze the IF heatmap of the best-performing multi-head attention mechanism across each data modality, as indicated in Tables \ref{tab:combined}. For instance, in the context of \textit{Speech Processing with WavLM-Large using DAAM}, Figure \ref{iemocap_gaam} shows a dense population of higher IF values at the lower layers, indicating these layers' significant role in modulating the input sequence. This suggests that foundational speech features are captured early on, with upper layers refining these features into more abstract representations. Conversely, \textit{Text Processing with Llama2-13B using GQDAAM}, illustrated in Figure \ref{llama2_gqgaam}, displays a more uniform distribution of IF across all layers, with a slight concentration in earlier layers. This pattern reflects a balanced approach to hierarchical feature extraction, where both lower and higher-level features are crucial, particularly those derived from the early to middle layers.

Similarly, \textit{Digital Image Processing with BEiT-Large using GQDAAM} in Figure \ref{cifar100_gqgaam} emphasizes the importance of lower layer features, which is consistent with the need for early-stage feature extraction in visual tasks, such as identifying edges and textures. These variations in IF distribution highlight the unique information processing needs of each modality. While speech and image processing rely heavily on primary feature extraction, text processing requires a combination of fundamental and more complex feature identification. The insights gained from IF analysis not only enhance the explainability of the models but also provide a quantifiable measure of feature significance, facilitating more informed decisions in model refinement and adaptation.

\begin{table}[H]
\centering
\footnotesize
\renewcommand{\arraystretch}{0.9} 
\setlength\tabcolsep{6pt}  
\scalebox{0.95}{  
\begin{tabular}{lcccccc}
    \toprule
    \multicolumn{7}{c}{\textbf{IEMOCAP 5-fold validation (F1-F5) using WavLM-Large}} \\
    \toprule
    \textbf{Method} & \textbf{F1} & \textbf{F2} & \textbf{F3} & \textbf{F4} & \textbf{F5} & \textbf{$\mu \pm \sigma$} \\
    \midrule
    LoRA+ (r=4)    & 27.6 & 25.7 & 31.7 & 25.1 & 16.8 & 25.4 $\pm$ 4.87 \\
    LoRA+ (r=8)    & 27.6 & 28.3 & 20.5 & 20.6 & 24.6 & 24.3 $\pm$ 3.32 \\
    LoRA (r=4)    & 49.9 & 51.5 & 58.2 & 52.6 & 52.7 & 53.0 $\pm$ 2.79 \\
    LoRA (r=8)    & 49.4 & 51.8 & 61.5 & 48.7 & 55.1 & 53.3 $\pm$ 4.66 \\
    MHA    & 62.7 & 59.9 & 61.7 & 61.3 & 65.7 & 62.3 $\pm$ 2.00 \\
    MHA $\rightarrow$ BN  & 62.7 & 59.9 & 62.9 & 64.8 & 66.6 & 63.4 $\pm$ 2.50 \\
    \textbf{DAAMv2}    & \textbf{66.1} & \textbf{60.0} & \textbf{66.3} & \textbf{65.2} & \textbf{65.4} & \textbf{64.6} $\pm$ \textbf{2.47} \\
    \textbf{GQDAAM}  & \textbf{66.5} & \textbf{65.4} & \textbf{68.7} & \textbf{65.9} & \textbf{66.8} & \textbf{66.7} $\pm$ \textbf{1.18} \\
    \textbf{DAAMv1}    & \textbf{67.2} & \textbf{64.6} & \textbf{68.1} & \textbf{67.9} & \textbf{69.0} & \textbf{67.4} $\pm$ \textbf{1.49} \\
    \midrule
    \multicolumn{7}{c}{\textbf{CIFAR100 5 Run validation (R1-R5) using BEiT-Large}} \\
    \toprule
    \textbf{Method} & \textbf{R1} & \textbf{R2} & \textbf{R3} & \textbf{R4} & \textbf{R5} & \textbf{$\mu \pm \sigma$} \\
    \midrule
    LoRA+ (r=4)    & 20.2 & 21.1 & 26.8 & 17.9 & 24.5 & 22.1 $\pm$ 3.17 \\
    LoRA+ (r=8)    & 25.0 & 32.9 & 22.9 & 29.1 & 27.5 & 27.5 $\pm$ 3.44 \\
    LoRA (r=4)    & 35.7 & 32.3 & 31.5 & 36.2 & 40.1 & 35.2 $\pm$ 3.08 \\
    LoRA (r=8)    & 38.1 & 40.0 & 42.3 & 41.6 & 39.6 & 40.3 $\pm$ 1.49 \\
    MHA    & 60.4 & 61.9 & 62.1 & 62.0 & 62.1 & 61.7 $\pm$ 0.75 \\
    MHA $\rightarrow$ BN  & 63.0 & 67.1 & 69.5 & 63.9 & 67.0 & 66.1 $\pm$ 2.25 \\
    \textbf{GQDAAM}  & \textbf{80.0} & \textbf{80.1} & \textbf{80.1} & \textbf{80.6} & \textbf{80.0} & \textbf{80.1} $\pm$ \textbf{0.24} \\
    \textbf{DAAMv1}  & \textbf{79.9} & \textbf{80.2} & \textbf{80.2} & \textbf{80.7} & \textbf{80.7} & \textbf{80.3} $\pm$ \textbf{0.32} \\
    \textbf{DAAMv2} & \textbf{80.2} & \textbf{80.4} & \textbf{81.0} & \textbf{80.3} & \textbf{81.0} & \textbf{80.6} $\pm$ \textbf{0.36} \\
    \midrule
    \multicolumn{7}{c}{\textbf{Modality Summary: Gaussian Parameters for Normalized Features}} \\
    \toprule
    \textbf{Modality} & \multicolumn{3}{c}{\textbf{Mean Offset}} & \multicolumn{3}{c}{\textbf{Scaled Variance}} \\
    \midrule
    Speech & \multicolumn{3}{c}{[-0.06, 0.10]} & \multicolumn{3}{c}{[1.88, 2.06]} \\
    Text   & \multicolumn{3}{c}{[-0.05, 0.07]} & \multicolumn{3}{c}{[1.94, 2.02]} \\
    Vision & \multicolumn{3}{c}{[-0.02, 0.02]} & \multicolumn{3}{c}{[1.98, 2.03]} \\
    \midrule
    \multicolumn{7}{c}{\textbf{AG News 3 Run validation (R1-R3) using Llama2-13B}} \\
    \toprule
    \textbf{Method} & \textbf{R1} & \textbf{R2} & \textbf{R3} & \multicolumn{3}{c}{\textbf{$\mu \pm \sigma$}} \\
    \midrule
    LoRA+ (r=4)    & 93.4 & 65.9 & 92.8 & \multicolumn{3}{c}{84.0 $\pm$ 12.8} \\
    LoRA+ (r=8)    & 95.0 & 69.8 & 94.6 & \multicolumn{3}{c}{86.5 $\pm$ 11.8} \\
    \textbf{DAAMv2}    & \textbf{94.4} & \textbf{94.5} & \textbf{94.6} & \multicolumn{3}{c}{\textbf{94.5} $\pm$ \textbf{0.08}} \\
    MHA $\rightarrow$ BN  & 94.5 & 94.5 & 94.7 & \multicolumn{3}{c}{94.6 $\pm$ 0.11} \\
    MHA    & 94.4 & 94.5 & 94.8 & \multicolumn{3}{c}{94.6 $\pm$ 0.16} \\
    \textbf{DAAMv1}    & \textbf{94.5} & \textbf{94.5} & \textbf{94.7} & \multicolumn{3}{c}{\textbf{94.6} $\pm$ \textbf{0.11}} \\
    LoRA (r=8)    & 94.9 & 94.6 & 94.9 & \multicolumn{3}{c}{94.8 $\pm$ 0.14} \\
    \textbf{GQDAAM}  & \textbf{94.8} & \textbf{94.9} & \textbf{94.9} & \multicolumn{3}{c}{\textbf{94.9} $\pm$ \textbf{0.06}} \\
    LoRA (r=4)    & 95.1 & 94.5 & 95.3 & \multicolumn{3}{c}{95.0 $\pm$ 0.30} \\
    \midrule
    \multicolumn{7}{c}{\textbf{High vs. Low IF Scores for IEMOCAP Validation using WavLM}} \\
    \toprule
    \textbf{Layer} & \textbf{F1} & \textbf{F2} & \textbf{F3} & \textbf{F4} & \textbf{F5} & \textbf{Avg.} \\
    \midrule
    9 (High) & \textbf{65.9} & \textbf{60.1} & \textbf{64.4} & \textbf{62.7} & \textbf{67.0} & \textbf{64.0} \\
    23 (Low) & 62.8 & 58.9 & 63.2 & 62.0 & 64.5 & 62.3 \\
    \midrule
    \multicolumn{7}{c}{\textbf{Accuracy for High and Low IF Layers using Llama2-13B and BEiT-Large}} \\
    \toprule
    \textbf{Dataset} & \multicolumn{3}{c}{\textbf{High IF Layers}} & \multicolumn{3}{c}{\textbf{Low IF Layers}} \\
    \midrule
    AGNews & \multicolumn{3}{c}{\textbf{94.9} (19-21)} & \multicolumn{3}{c}{94.7 (37-39)} \\
    CIFAR100 & \multicolumn{3}{c}{\textbf{72.6} (10-12)} & \multicolumn{3}{c}{64.7 (22-24)} \\
    \bottomrule
\end{tabular}
}
\vspace{2mm}
\caption{Comparison of results from IEMOCAP, CIFAR100, modality summary, and AG News datasets, including IF layers for different methods. Best performance is indicated in bold.}
\label{tab:combined}
\end{table}

\begin{figure*}[t]
\centering
\subfloat[Speech Processing with \textit{WavLM-Large} using DAAM.]{
  \includegraphics[width=0.31\textwidth, height=3.5cm]{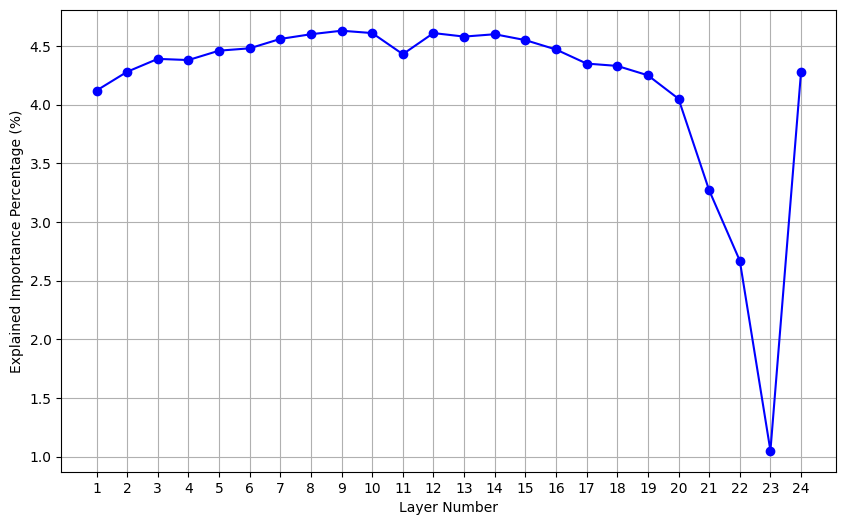}
  \label{wavlm_explained}
}\hspace{1mm}
\subfloat[Text Processing with \textit{Llama2-13B} using DAAM.]{
  \includegraphics[width=0.31\textwidth, height=3.5cm]{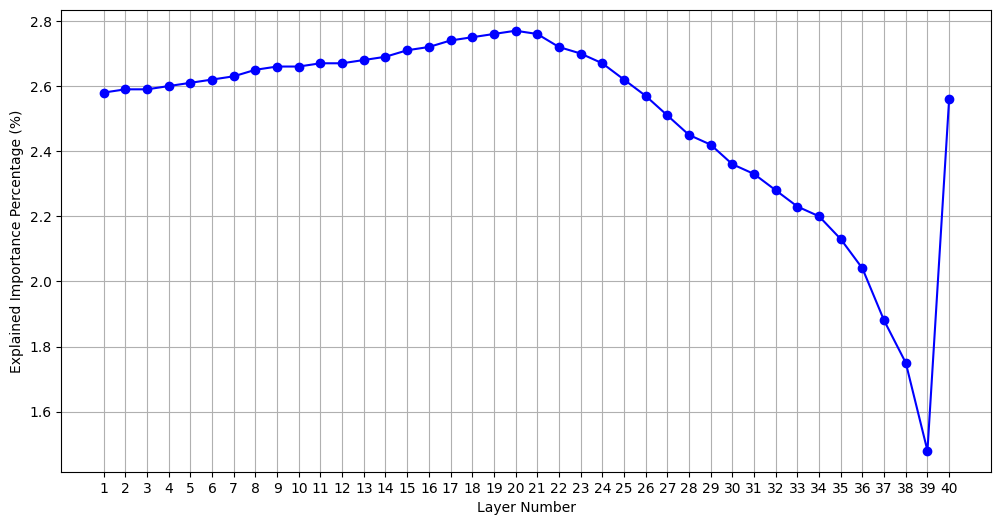}
  \label{llama_explained}
}\hspace{1mm}
\subfloat[Image Processing with \textit{BEiT-Large} using DAAM.]{
  \includegraphics[width=0.31\textwidth, height=3.5cm]{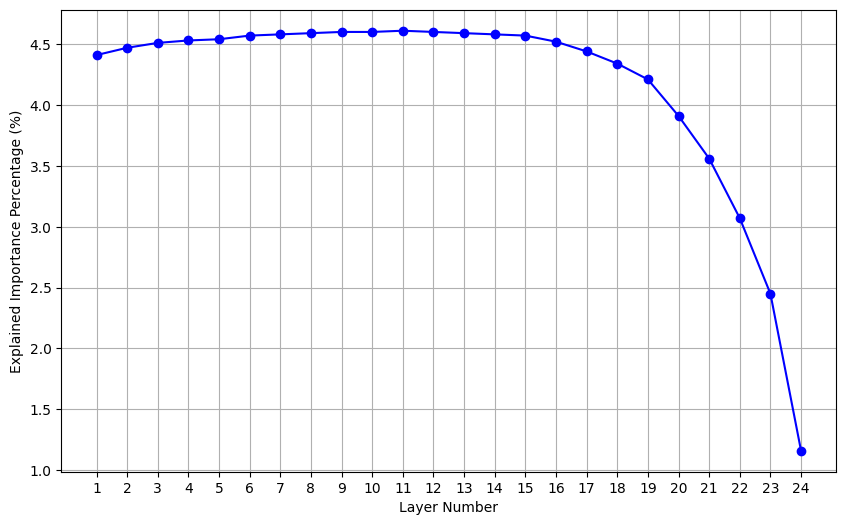}
  \label{beit_explained}
}
\caption{Percentage contribution of each layer to attention weights in different downstream tasks (for best performing DAAM-based models using $g : 8$).}
\label{contribute}
\end{figure*}
\subsection{Ablation Studies}
We validate that the IF from DAAM and GQDAAM attention weights accurately identifies key feature extraction regions affecting model performance. This is achieved by reassessing experiments focused on layers with low and high IF scores, aiming to understand the link between IF scores and the significance of the highlighted features.

\noindent Across Speech, Text, and Vision modalities, higher IF scores consistently align with improved model performance and vice versa. 
For the Speech downstream task, High IF layer (Layer 9) consistently outperforms low IF layer (Layer 23) across all folds (Table \ref{tab:combined}).
For the Text and Vision downstream tasks (Table \ref{tab:combined}), layers with higher IF achieve better performance, especially in Vision. In MHA, attention weights primarily indicate the level of correlation between different parts of the input sequence \cite{Lovisotto2022GiveMY}. Each element's weight reflects its relevance to every other element within the same sequence. However, this approach does not directly translate to the performance on downstream tasks. For instance, the authors in \cite{ioannides23_interspeech} derive normalized self-attention weights for SER on IEMOCAP using WavLM-Large, identifying layer 23 as pivotal. While useful for their use case, this only indicates inter-layer correlation, and not a direct link to better or worse performance and would be misleading to use these weights for that use case (as shown in Table \ref{tab:combined} using DAAM for the same task). In contrast, DAAM-based learning dynamically adjusts attention weights tailoring attention to improve feature representation aligned with the model's end goal. Analysis of Figure \ref{contribute} indicates earlier layers, exhibit more meaningful features, and contribute more to model performance, suggesting potential overparameterization in later layers \cite{zhang2022platon}.
\section{Limitations \& Future Work}
DAAM's fixed number of Gaussians can limit its adaptability across different datasets and tasks. This can be improved by adopting a Bayesian approach to dynamically select the optimal number of Gaussians. Utilizing criteria like the Bayesian Information Criterion (BIC) or a neural network to predict the number based on input characteristics can enhance performance and efficiency, allowing the model to better adapt to varying data distributions and complexities. Future work should explore DAAM in additional tasks, datasets, grounding experiments \cite{grounding}, and beyond feature extraction, including model compression using attention weights during training (crucial for resource-limited applications) \cite{back2023magnitude}.
\section{Conclusion}
\label{conclusion}
In this work, we introduce Multi-Head DAAM and the Density Adaptive Transformer. We demonstrate their effectiveness in enhancing model performance, particularly with highly non-stationary data such as Speech and Vision. Results show that combining learnable mean and variance for every Gaussian Distribution enables dynamic feature significance recalibration and approximation of any Probability Distribution across multiple modalities. 
Combining this mechanism with the dot-product attention mechanism enhances performance with a minimal increase in parameters (0.016\%-0.08\% compared to GQA models) and at least 44\% less total parameters than LoRA. Finally, we introduce the Importance Factor for improved model explainability.

\newpage
\bibliographystyle{IEEEtran}
\bibliography{neurips_2024.bib}

%%%%%%%%%%%%%%%%%%%%%%%%%%%%%%%%%%%%%%%%%%%%%%%%%%%%%%%%%%%%

\newpage
\appendix
\section{Reproducibility}
The source code has been provided in the following GitHub repository: \url{https://github.com/gioannides/DAAM-PEFT-paper-code}
\section{Multi-Head Mixture of Densities Adaptive Attention Mechanism Extension}
\label{optimized}
This section presents an extension of the Multi-Head Density Adaptive Attention Mechanism (DAAM), focusing on enhancing the stability of the training process and the model's efficiency by \textbf{significantly} reducing the number of learnable parameters even further. The proposed method integrates multi-head attention mechanisms with Gaussian mixtures and skip connections to provide a more refined and adaptable approach to handling complex datasets.

The extended DAAM incorporates multiple attention heads, each with its Gaussian mixture model, to process different segments of the input tensor in parallel. This approach allows for a more diverse and comprehensive understanding of the data, leading to increased model robustness and efficiency.

Additionally, as illustrated in Algorithm \ref{alg:density_block} then add the original input features ($X$) to the augmented one ($X'$) for enhanced stability during training (i.e. $X' \gets X' + X$).

\subsection{Algorithmic Details}
This algorithm which forms the core of the extended DAAM, implementing the Gaussian mixture model within each attention head can be found in Algorithm \ref{alg:attention_modified_tensor}. Key elements include initialization of Gaussian parameters and mean offsets, and a forward pass handling.
\begin{algorithm}
\caption{Mixture of Densities Adaptive Attention Mechanism}
\label{alg:attention_modified_tensor}
\begin{algorithmic}[1]
\REQUIRE Input tensor \( \mathbf{x} \), norm axis \( \text{normAxis} \), \( N \) Gaussians, \( \epsilon \)
\ENSURE Attention-modified \( \mathbf{x} \)
\STATE Initialize \( \mathbf{m} \), \( \mathbf{c} \) of size \( N \), \( \boldsymbol{\mu} \gets \text{mean}(\mathbf{x}, \text{axis}=\text{normAxis}) \), \( \boldsymbol{\sigma}^2 \gets \text{var}(\mathbf{x}, \text{axis}=\text{normAxis}) + \epsilon \), \( \text{mixture} \gets 1 \)
\FOR{$i = 0$ to $N-1$}
    \STATE \( \boldsymbol{\mu}_i^{\text{adj}} \gets \boldsymbol{\mu} + \mathbf{m}[i] \)
    \STATE \( \mathbf{y}_i \gets \frac{\mathbf{x} - \boldsymbol{\mu}_i^{\text{adj}}}{\sqrt{\boldsymbol{\sigma}^2}} \)
    \STATE \( \mathbf{g}_i \gets \frac{\exp(-\frac{\mathbf{y}_i^2}{2\mathbf{c}[i]^2})}{\sqrt{2\pi\mathbf{c}[i]^2}} \)
    \STATE \( \text{mixture} \gets \text{mixture} \cdot \mathbf{g}_i \)
\ENDFOR
\STATE Normalize \( \text{mixture} \) across \( \text{normAxis} \)
\STATE \( \mathbf{x'} \gets \mathbf{x} \cdot \text{mixture} \)
\RETURN \( \mathbf{x'} \)
\end{algorithmic}
\end{algorithm}

\begin{algorithm}
\caption{Multi-Head Density Adaptive Attention}
\label{alg:multi_head_gaussian_adaptive_attention}
\begin{algorithmic}[1]
\REQUIRE $x$ (input tensor), normDimSize, numHeads, normAxis, c
\ENSURE Concatenated attention output tensor
\STATE Initialize an array $\mathbf{attentionHeads}$ of size numHeads
\FOR{$i = 1$ to numHeads}
    \STATE $\mathbf{attentionHeads}[i] \gets \text{GAAM}($\textbf{normDimSize}, \textbf{normAxis}, \textbf{c}, \textbf{eps}$)$
\ENDFOR
\STATE Initialize an empty list $\mathbf{outputs}$
\FOR{each $\mathbf{head}$ in $\mathbf{attentionHeads}$}
    \STATE $\mathbf{headOutput} \gets \text{apply}(\mathbf{head}, x)$
    \STATE Append $\mathbf{headOutput}$ to $\mathbf{outputs}$
\ENDFOR
\STATE $\mathbf{output} \gets \text{concatenate}(\mathbf{outputs}, \text{dim}=\text{normAxis})$
\RETURN $\mathbf{output}$
\end{algorithmic}
\end{algorithm}

\begin{algorithm}[H]
\caption{Density Block}
\label{alg:density_block}
\begin{algorithmic}[1]
\REQUIRE $x$ (input tensor), normAxes, numHeads, numGaussians, paddingValue, eps
\ENSURE Final modified tensor
\FOR{each layer in MultiHeadDensityAdaptiveAttention}
    \STATE $x$ $\gets$ layer($x$) + $x$
\ENDFOR
\RETURN $x$
\end{algorithmic}
\end{algorithm}

\subsection*{Extended Results}
We repeat all experiments previously carried out but with the Mixture of DAAM instead (see Table \ref{tab:combined}). It is evident that Mixture of DAAM not only outperforms DAAM but it also reduces its overall trainable parameter count significantly (see Table \ref{tab:combined} inside the parentheses where only the parameters relevant to the attention mechanism are provided).
\begin{table}[H]
\centering
\footnotesize % Smaller font size for the table content
\begin{tabular}{lccc}
\toprule
\textbf{Mechanism} & \textbf{Heads} & \textbf{$N$ Gaussians} & \textbf{Parameters} \\
\midrule
Mixture of DAAM & \(g: 8\) & 4 & 64 \\
\bottomrule
\end{tabular}
\vspace{2mm}
\caption{Number of learnable parameters for Mixture of Densities Aadaptive Attention Mechanism.}
\label{tab:attention_comparison2}
\end{table}

\begin{table}[H]
\centering
\footnotesize  % Sets a smaller font size
\begin{tabular}{lcccccc}  % Ensure there are six data columns plus one for the method names
    \toprule
    \textbf{Method} & \textbf{F1} & \textbf{F2} & \textbf{F3} & \textbf{F4} & \textbf{F5} & \textbf{$\mu$ $\pm$ $\sigma$} \\
    \midrule
    Mixture of DAAM (with 2 \textit{conv.} layers)    & 67.8 & 69.5 & 65.1 & 68.7 & 67.8 & 67.9 $\pm$ 1.35 \\
    \bottomrule
\end{tabular}
\vspace{2mm}
\caption{IEMOCAP 5-fold validation (F1-F5) using WavLM-Large.}
\label{tab:iemocap2}
\end{table}

\begin{table}[H]
\centering
\footnotesize  % Sets a smaller font size
\begin{tabular}{lcccccc}  % Ensure there are six data columns plus one for the method names
    \toprule
    \textbf{Method} & \textbf{R1} & \textbf{R2} & \textbf{R3} & \textbf{R4} & \textbf{R5} & \textbf{$\mu$ $\pm$ $\sigma$} \\
    \midrule
    Mixture of DAAM (with 2 \textit{conv.} layers)   & 80.6 & 79.7 & 80.5 & 80.3 & 80.3 & 80.3 $\pm$ 0.3 \\
    \bottomrule
\end{tabular}
\vspace{2mm}
\caption{CIFAR100 5 Run validation (R1-R5) using BEiT-Large.}
\label{tab:cifar100-2}
\end{table}

\begin{table}[H]
\centering
\begin{tabular}{lcccc}
    \toprule
    \textbf{Method} & \textbf{R1} & \textbf{R2} & \textbf{R3} & \textbf{$\mu$ $\pm$ $\sigma$} \\
    \midrule
    Mixture of DAAM (with 2 \textit{conv.} layers)   & 94.5 & 94.6 & 94.6 & 94.6 $\pm$ 0.05 \\
    \bottomrule
\end{tabular}
\vspace{2mm}
\caption{AG News 3 Run validation (R1-R3) using Llama2-13B.}
\label{tab:agnews2}
\end{table}

\end{document}